# Adaptive kNN Using Expected Accuracy for Classification of Geo-Spatial Data

Mark Kibanov
University of Kassel, ITeG Center
Knowledge and Data Engineering
Wilhelmshöher Allee 73
34121, Kassel
kibanov@cs.uni-kassel.de

Martin Becker
University of Würzburg
DMIR Group
Am Hubland
97074, Würzburg
becker@informatik.uni-wuerzburg.de

Juergen Mueller
University of Kassel, ITeG Center
Knowledge and Data Engineering
Wilhelmshöher Allee 73
34121, Kassel
mueller@cs.uni-kassel.de

Martin Atzmueller
Tilburg University (CSAI)
Warandelaan 2
5037 AB, Tilburg
m.atzmuller@uvt.nl

Andreas Hotho
University of Würzburg
DMIR Group
Am Hubland
97074, Würzburg
hotho@informatik.uni-wuerzburg.de

Gerd Stumme
University of Kassel, ITeG Center
Knowledge and Data Engineering
Wilhelmshöher Allee 73
34121, Kassel
stumme@cs.uni-kassel.de

## ABSTRACT

The $k$-Nearest Neighbor (kNN) classification approach is conceptually simple – yet widely applied since it often performs well in practical applications. However, using a global constant $k$ does not always provide an optimal solution, e. g., for datasets with an irregular density distribution of data points. This paper proposes an adaptive kNN classifier where $k$ is chosen dynamically for each instance (point) to be classified, such that the expected accuracy of classification is maximized. We define the expected accuracy as the accuracy of a set of structurally similar observations. An arbitrary similarity function can be used to find these observations. We introduce and evaluate different similarity functions. For the evaluation, we use five different classification tasks based on geo-spatial data. Each classification task consists of (tens of) thousands of items. We demonstrate, that the presented expected accuracy measures can be a good estimator for kNN performance, and the proposed adaptive kNN classifier outperforms common kNN and previously introduced adaptive kNN algorithms. Also, we show that the range of considered $k$ can be significantly reduced to speed up the algorithm without negative influence on classification accuracy.

## CCS CONCEPTS

• **Computing methodologies → Instance-based learning**;

## KEYWORDS

Classification, kNN, Geo-spatial Data, Adaptive Algorithms

## 1 INTRODUCTION

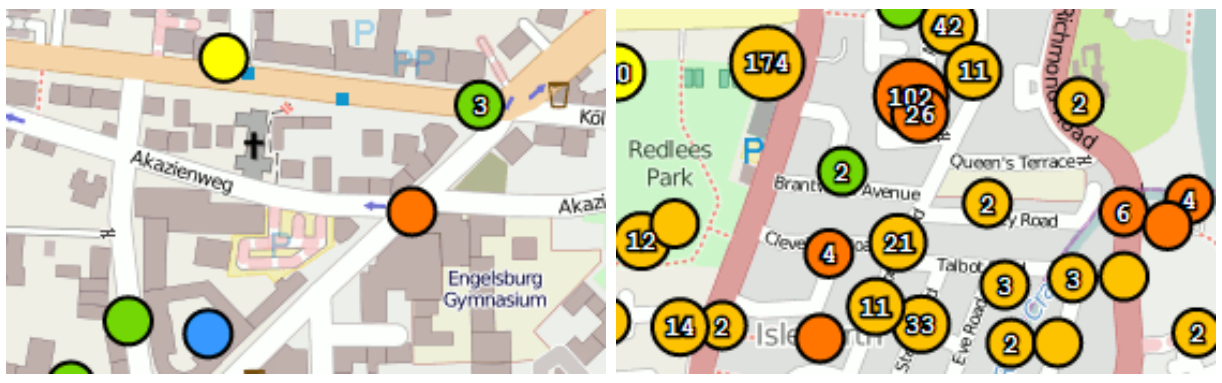

**(a) Sparse coverage.** **(b) Dense coverage.**

**Figure 1: Examples of different distributions of data points in the WideNoise Dataset. Different colors denote different noise levels.**

K-nearest neighbors is a simple, yet effective classification algorithm. For example, one of its big advantages is that classification results can be easily interpreted. Consequently, it has been applied in a large variety of settings . The basic assumption of the standard kNN is that a data point (instance) is often a member of the same class as the majority of its $k$ nearest neighbors, where $k$ is fixed for all data points to classify. However, many datasets have an irregular distributions of data points. Consider Figure 1, for example, where areas with different density of points are shown. Due to the density variations, it would be useful to consider different $k$ for different points (items): e. g., a simple intuition would be to use more neighbors in dense areas and less in the areas with sparse neighbors. However, if large $k$ is used, it can be hardly influenced by the largest class; if $k$ is too small, the results can be strongly influenced by a small number of neighbors, e. g., noisy data. Thus, finding an optimal value for $k$ can be a challenging task. In this paper, we address this issue and propose a general approach for an adaptive kNN classifier, called *ACKER – **A**daptive **C**lassifier based on **KNN** **E**xpected accu**R**acy*).

That is, the proposed ACKER algorithm aims to find an optimal $k$ for each point that should be classified. In these terms, an optimal $k$ is defined in such way, that expected accuracy of standard kNN applied to a given point with this $k$ is maximal. To determine the expected accuracy for a point and its $k$ nearest neighbors, we search *similar* points (in the training dataset) and then compute the accuracy of the kNN classification for them. The set of most similar points can be determined using similarity function. We present different similarity functions and further parameters of ACKER in Section 4 and discuss their influence on the performance of the algorithm in Section 5.

In this paper, we concentrate on geo-spatial data to illustrate our approach. For evaluation, we use five different classification tasks, constructed from three datasets with geo-spatial data. We demonstrate the effectiveness of our proposed expected accuracy measure for the algorithm's predictive performance, and show that the ACKER algorithm outperforms standard kNN and another, previously introduced version of an adaptive kNN algorithm. Finally, we discuss how the number of considered candidate values of $k$ can be reduced to improve the runtime of ACKER.

The contributions of this paper can be summarized as follows:

(1) We propose a measure for estimating of kNN performance, i.e., the *expected accuracy* and show its effectiveness.
(2) We introduce ACKER and show, that its performance is higher than that of standard kNN and previously introduced version of adaptive kNN.

The ACKER algorithm was never introduced before. It is the first time, we explain this algorithm and evaluate it. We use geo-spatial data for the evaluation. This choice is discussed and motivated in Section 3.3. However, we plan to test this approach with more diverse high dimensional data.

The rest of the paper is structured as follows. Section 2 discusses related work. Then, we recall some basic concepts in Section 3. In Section 4 we provide a description of the ACKER algorithm and related terms, such as *expected accuracy*, and *similarity functions*. The datasets and the results of the evaluation are described in Section 5. Finally, Section 6 concludes the paper with a summary and interesting options for future work.

## 2 RELATED WORK

The idea of kNN was proposed by Fix and Hodges [5]. Sun and Huang [12] present an adaptive kNN algorithm which aims to select the optimal $k$ for the classification. Kamdar Rajani [6] presented an implementation of this adaptive kNN approach for the MapReduce paradigm. This approach chooses an optimal $k$ based on an optimal $k$ for $l$ nearest neighbors. We compare the proposed adaptive algorithm with ACKER for the presented geo-spatial data. We do not use a MapReduce approach as we make our experiments using 2-dimensional data and thus the runtime is not a critical issue in the presented tasks.

Another adaptive kNN approach uses confidence measure to determine an optimal $k$ [10]. Confidence is defined as a share of neighbor items which belong to the majority class. We do not consider this algorithm in our case – because if possible range of values of $k$ is $\{1..n\}$, then maximal confidence (100%) will be always reached, if $k = 1$. In this case, either this algorithm is equivalent to standard kNN, or it is necessary to remove some values from *Range*, e. g., 1. However, in considered datasets, $k = 1$ was an optimal $k$ for most cases. Thus, we do not use this algorithm as a baseline in this paper.

Wang et al. [13, 14] proposed a simple adaptive distance measure for improving the nearest neighbor identification. Ougiaroglou et al. [9] focuses on improving the performance of kNN algorithms by using an adaptive number of nearest neighbors in order to reduce the runtime complexity of the algorithm. In contrast to that, the ACKER concentrates on improving of overall accuracy. In the Section 5.4 we show how to improve the runtime of the ACKER.

For the classification, using uncertain information, techniques for reliability-based classification have been developed. Senge et al. [11], for example, present an approach for binary classification that produces a prediction together with a quantification of its reliability. Similar to that approach, the ACKER classifier derives an expected accuracy for classification. Using this parameter, the presented approach can then be tuned in order to provide a prediction that is as exact as possible w.r.t. the given information used for building the model, and for obtaining different classes of the prediction space in terms of the expected accuracy.

Many applications can utilize a more accurate kNN classification: e. g., Fang et al. [3] proposed an algorithm for effective join algorithm for trajectory data. That algorithm enables finding nearest Combination of this approach together with our kNN approach may enable new application and algorithms, e. g., for classification of moving objects.

## 3 BACKGROUND

In this Section, we define the general classification problem, the standard kNN algorithm, and two measures for evaluating the classification quality.

### 3.1 Classification and kNN Algorithm

We consider the problem of classification. That is, we assume that each point $p \in P$, from a set of points $P$, can be assigned to a class $x \in X$, from a finite set of classes $X$, by a classification function $c\colon P \rightarrow X$. The set of all classified points is $C = \{(p, x) \in P \times X \mid x = c(p)\}$. In general, the classification function $c$ is unknown. Thus, given a training dataset – a subset of all classified points $C_{\text{train}} \subset C$ – the goal is to approximate $c$ by a classification function $\hat{c}$.

The kNN algorithm results in an approximation $\hat{c}_k$ of the classification function $c$ which works as follows: Given a distance function $dist\colon P \times P \rightarrow \mathbb{R}$, the set $N_{p,k} \subseteq T$ denotes the $k$ nearest neighbors of $p$ in training set $C_{\text{train}}$. We call the pair $(p, N_{p,k})$ the $k$-environment of $p$ (see Section 4.2). The class $\hat{c}_k(p)$ assigned to point $p$ is the majority class within this set of $k$ nearest neighbors $N_{p,k}$. In our particular case we will use geo-spatial points defined by longitude and latitude, i.e., $P = [-180; 180] \times [-90; 90]$, and the Euclidean distance function $d$ which approximates the great circle distance well enough for our purposes.

## 3.2 Classification Quality Measures

Given an estimated classification function $\hat{c}$ and a testing set $C_{test} \subset C, C_{test} \cap C_{train} = \emptyset$, we use the accuracy ($acc$) as a measure of classification quality. Accuracy is defined as the ratio of correctly classified data points $C_{test} \cap C^{\hat{c}}_{test}$ versus the overall number of data points being classified $C_{test}$, where $C^{\hat{c}}_{test} = \{(p, \hat{c}(p)) \mid (p, x) \in C_{test}\}$ is the set of data points classified by the estimated classification function $\hat{c}$:

$$acc(\hat{c}, C_{test}) = \frac{|C_{test} \cap C^{\hat{c}}_{test}|}{|C_{test}|} \quad (1)$$

Given, a function $score_{\hat{c}}: P \rightarrow \mathbb{R}$ that shows a probability or a score of classification correctness for $\hat{c}(p)$. We use the Receiver operating characteristic Area Under Curve (ROC AUC) function as measure of the quality of the score function[4].

*ROC AUC* is defined as follows. Function *score* predicts, how likely given classification result $\hat{c}(p)$ is correct. We use *expected accuracy* as a *score* function (see Section 4.2). $C^t$ denotes predictions with *score* higher than given threshold $t$: $(p, \hat{c}) \in C^t$, iff $score(p, \hat{c}) > t$. The True Positive Rate function, $TPR(t)$, shows which portion of predictions $\hat{c}(p)$ with $score(p, \hat{c}) > t$ is correct: $TPR(t) = \frac{|C_{test} \cap C^{\hat{c}}_{test} \cap C^t|}{|C_{test} \cap C^{\hat{c}}_{test}|}$. The False Positive Rate Function, $TPR(t)$, shows which portion of predictions $\hat{c}(p)$ with $score(p, \hat{c}) > t$ is false: $FPR(t) = \frac{|(C^{\hat{c}}_{test} \setminus C_{test}) \cap C^t|}{|C^{\hat{c}}_{test} \setminus C_{test}|}$. Then the *Receiver operating characteristic, ROC*, curve characterizes *score* function and is defined as follows: $ROC(x) = y$, iff $x = FPR(t)$ and $y = TPR(t)$. ROC Area Under the Curve (ROC AUC) measure quantifies the quality of *score*: $ROC\ AUC = \int_{-\infty}^{\infty} TPR(T)(-FPR'(T))dT$.

## 3.3 Discussion

*kNN* is a basic and well-known classification algorithm that has a number of advantages [8]. First, *kNN* has no training time, as it is a lazy learning method. *kNN* can be used as an incremental learner. For some applications, it is important that kNN is easily interpretable. On another side, *kNN* has disadvantages, some of which are addressed by ACKER, particularly, classification accuracy and tolerance to noise. Further disadvantages, such as the fact that the *kNN* classification uses more storage space than other algorithms, has low speed of classification and is intolerant to irrelevant attributes, are not critical for the two-dimensional geo-spatial data.

# 4 ACKER – ADAPTIVE CLASSIFIER BASED ON K NEAREST NEIGHBOUR EXPECTED ACCURACY

The main idea of the ACKER algorithm is to find optimal $k$ for each point (instance) and to use this $k$ with standard kNN, (see Section 4.1). Given a point $p$, the algorithm estimates the *expected accuracy* for different possible values of $k$, and chooses a $k$ that provides a maximum value of the *expected accuracy*, (see Section 4.2). We apply kNN regression to $k$-environments of points to estimate the *expected accuracy*. We use different *similarity functions* to find the most similar (nearest) points with its $k$-environments for given point $p$ and its $k$-environments (see Section 4.3).

In this section we describe the proposed classification algorithm and show how the *expected accuracy* can be computed using different *similarity functions*. Finally, we discuss the runtime complexity of the algorithm.

## 4.1 Adaptive k-Nearest Neighbor

We propose an adaptive kNN algorithm – **ACKER**: Adaptive Classifier based on kNN Expected accuRacy. As mentioned above, the main idea of ACKER is to choose $k$ dynamically for each point to maximize expected accuracy (see Section 4.2). In other words, we need to find $k$ so that $exp_acc(p, k) = max\{exp_acc(p, k) | k \in Range\}$, where *Range* is defined as the range of possible values for $k$. The standard kNN classifier with optimal $k$ is applied afterwards.

The ACKER algorithm (see Listing 1) has four input variables:

- $p$ is a point for which the classification should be made;
- $Range \subseteq \mathbb{N}$ defines the set of candidates $k$.
- *SimilarityFunction* defines which similarity function should be used for estimation of the expected accuracy.
- $l$ is the parameter for the expected accuracy estimation (see Equation 2).

In addition, the following functions (indicated by teletype font) are used in the algorithm:

- `expected_accuracy`($p$, $k$, *SimilarityFunction*, $l$) returns the expected accuracy of kNN for point $p$ and its $k$-environment (see Listing 2);
- `kNN`($p$, $k$) returns the result of kNN algorithm for the point $p$ based on $k$ neighbors.

**Algorithm 1:** ACKER

**Data:** point $p$, set *Range*, function $f$, integer $l$
**Result:** Class of point $p$

1 $optimal_k = 0$
2 $max_accuracy = -1$
3 **for** *k in Range* **do**
4     $candidate_accuracy$ = `expected_accuracy`$(p, k, f, l)$
5     **if** $candidate_accuracy > max_accuracy$ **then**
6         $optimal_k = k$
7         $max_accuracy = candidate_accuracy$
8     **end**
9 **end**
10 $predicted_class$ = `kNN`$(p, optimal_k)$
11 return ($predicted_class$)

The workflow of the proposed algorithm, Listing 1, can be described, as follows. First, the values of $optimal_k$ and $max_accuracy$ are initialized. The variable $optimal_k$ represents the value of $k$ with the maximal expected accuracy for $p$ found till current iteration; and $max_accuracy$ is the value of that maximal expected accuracy. These variables are initialized with values so, that they are rewritten in the first iteration. The algorithm iterates over all values of $k \in Range$ (lines 3 – 9). For each $k$ the algorithms computes an expected accuracy $candidate_accuracy$ (line 4) and compares it with the maximum expected accuracy found previously (lines 5 – 8). The algorithm chooses the $k$ with the highest expected accuracy and applies the standard kNN algorithm in order to classify the point (line 10).

Next, we discuss, how the *expected accuracy* can be computed using various *similarity functions*.

### 4.2 Expected Accuracy Measure

The intuition behind the *expected accuracy* can be described as follows. Given a point $p$ with its $k$-environment, find a set of points $C_{sim,l}(p,k) = \{p'_1 ... p'_l\} \subseteq C_{train}$ with similar $k$-environments. As the $k$-environments are similar, we assume, the accuracy of the kNN-based classification function for given point $p$, $acc(\hat{c}_k(p), \{p\})$, correlates with the accuracy of kNN-based classification function for $C_{sim,l}(p,k)$. So, the expected accuracy is defined as follows:

$$exp_acc_l(p,k) = acc(\hat{c}_k, C_{sim,l}(p,k)) \quad (2)$$

Here, we suppose that different regions are statistically similar. Sure, for some data it may be not the case, but in all considered datasets we found that expected accuracy is a good indicator for a real accuracy.

Note, that it is necessary to know the correct class for all points $p' \in C_{sim,l}(p,k)$, i.e. $C_{sim,l}(p,k) \subseteq C_{train}$. Listing 2 shows, how the expected accuracy (see Equation 2) can be computed.

**Algorithm 2:** Expected Accuracy

**Data:** point $p$, integer $k$, function $f$, integer $l$
**Result:** Expected Accuracy for `kNN(p, k)`

1 Find set $C_{sim,l}(p,k)$ of $l$ different points $p'$ with minimal difference to reference function w.r.t. $f$:
$sim(p,p') = |f(p,k) - f(p',k)|$,
$C_{sim,l}(p,k) = argmax(\sum sim(p,p',k))$(see Equation 4)
2 $correctly_predicted$ = number of correct predictions of `kNN`$(p',k)$, where $p' \in C_{sim,l}(p,k)$
3 $expected_accuracy = \frac{correctly_predicted}{l}$
4 return ($expected_accuracy$)

Next, we discuss which functions (features of $k$-environments) can be used to find $C_{sim,l}(p,k)$.

### 4.3 Similarity Function

The *similarity function* $sim\colon P \times P \times \mathbb{N} \to \mathbb{R}$ defines, how similar $k$-environment of two given points are. In some cases, similarity is not dependent on $k$. We then simply write $sim\colon P \times P \to \mathbb{R}$. Consider a function $f : P \times \mathbb{N} \to \mathbb{R}$ that characterizes a point and its $k$-environment. Then, a *similarity function* based on $f$ can be defined as:

$$\begin{aligned} sim_f &: P \times P \times \mathbb{N} \to \mathbb{R} \\ &(p,p',k) \mapsto -dist(f(p,k), f(p',k)), \end{aligned} \quad (3)$$

where $dist(x,y)$ denotes an Euclidean distance. The more similar two points with their $k$-environments are, the **higher is the value of** $sim_f$. Note, that the function $f$ can also characterize points and it neighbors by tuple, e. g., $f : P \times \mathbb{N} \to \mathbb{R} \times \mathbb{R}$, see Equation 9, this does not change the definition of $sim_f$.

$C_{sim,l}(p,k)$ is defined as the set containing the $l$ points whose $k$-environments are most similar to the one of $p$ with respect to the given similarity function, also written as

$$C_{sim,l}(p,k) = argmax^l_{p'}(sim(p,p',k)), \quad (4)$$

here and further $p, p' \in P$.

We introduce three similarity functions $sim_f$. Furthermore, we show, how the previously introduced adaptive kNN [12] can be described using our framework (particularly with a special $sim_f$).

$sim_{avg_dist}$: We denote by $avg_dist(p,k)$ the average distance to the $k$ nearest neighbors of point $p$:

$$avg_dist(p,k) = \frac{\sum_{i=1}^{k} dist(p,p_i)}{k} \quad , \quad (5)$$

here and in the sequel, $p_i$ is defined as the $i$th nearest neighbor to $p$. The similarity function based on $avg_dist$ is defined as follows:

$$sim_{avg_dist}(p,p',k) = -dist(avg_dist(p,k), avg_dist(p',k)) \quad (6)$$

$sim_{max_dist}$: We denote by $max_dist(p,k)$ the distance to the $k$th nearest neighbor of the point $p$:

$$max_dist(p,k) = dist(p,p_k) \quad (7)$$

The similarity function for two points $p$ and $p'$ based on $max_dist$ is defined as follows:

$$sim_{max_dist}(p,p',k) = -dist(max_dist(p,k), max_dist(p',k)) \quad (8)$$

$sim_{max_avg_comb}$: We denote by $max_avg_comb(p,k)$ a function that returns a tuple of results of $avg_dist$ (s. Equation 5) and $max_dist$ (s. Equation 7):

$$max_avg_comb(p,k) = (max_dist(p,k), avg_dist(p,k)) \quad (9)$$

The similarity function based on $max_avg_comb$ is defined as follows:

$$\begin{aligned} &sim_{max_avg_comb}(p,p',k) \\ &\quad = -dist(max_avg_comb(p,k), max_avg_comb(p',k)) \end{aligned} \quad (10)$$

$sim_{lat_lon}$: Sun and Huang [12] introduced an adaptive kNN classifier, where they proposed to choose k which is optimal for $l$ nearest neighbors of point $p$. We use this approach as a baseline for the evaluation (see Section 5). We set $f$ as follows in order to write adaptive kNN introduced in [12] in ACKER notation:

$$lat_lon(p) = (lat(p), lon(p)) \quad (11)$$

Here, $lat_lon(p)$ are the coordinates (latitude and longitude) of the point. The similarity function based on $lat_lon$ is defined as the Euclidean distance between point and its neighbors:

$$sim_{lat_lon}(p,p') = -dist(lat_lon(p), lat_lon(p')) \quad (12)$$

This example shows, how in some cases the ACKER approach can be considered as generalization of other adaptive kNN algorithms. in order to do so, it is necessary to adjust similarity function.

The influence of choice of similarity function on overall classification accuracy is shown in the Section 5.3. Next, we discuss the runtime complexity of ACKER and particularly for computing $C_{sim,l}$.

## 4.4 Complexity

We show the runtime complexity of computing $C_{sim,l}$ (and expected accuracy) for given point and its $k$-environment. Based on that, can we estimate the runtime of the ACKER algorithm.

Complexity of finding $C_{sim,l}(p,k)$: If the similarity function is based on function $f: P \times \mathbb{N} \to \mathbb{R}$, the values of $f(p), p \in C_{train}$ can be precomputed offline and indexed using a B+ tree. Given the value of $f(p)$, we need to find $C_{sim,l}(p,k)$. Search in a B+ tree can be done in $O(\log n)$, where $n$ is the number of elements of the tree and thus $O(\log |C_{\text{train}}|)$. The leaves of B+ tree can be linked to each other, so it is necessary to find only one element (which has the nearest value to $f(p)$) and consider the neighbor elements from it. Thus, the complexity of computing the expected accuracy in this case is $O(\log |C_{\text{train}}|)$.
If $f: \ldots \to \mathbb{R} \times \ldots \times \mathbb{R}$, the $kd$−Tree should be used instead of B+ tree to index values of $f(p), p \in P$. The search of one element is performed in $O(\log n) = O(\log |C_{\text{train}}|)$. The elements in $kd$−trees are not linked, so it is necessary to search for $l$ elements, this can be performed in $O(l \times \log |C_{\text{train}}|)$.
All steps in Listing 2, except the first line (finding $C_{sim,l}$), can be done in $O(1)$. Thus, computing expected accuracy is the same as finding $C_{sim,l}(p,k)$.

Complexity of ACKER: In the **for**–loop (lines 3 – 9 in Listing 1) the expected accuracy should be computed for each $k \in Range$, that means $O(|Range| \times \log |C_{\text{train}}|)$ or $O(|Range| \times l \times \log |C_{\text{train}}|)$ complexity for the ACKER algorithm, depending on similarity function, as described above.
Additionally, it is required to compute values of $f(p,k)$ for all $k \in Range$, where $p$ is the point that should be classified. In worst case, it is necessary to know all $k$ nearest neighbors to compute $f(p,k)$. So, it is needed to find $\max(Range)$ nearest neighbors. If the points in training set are indexed using $kd$−tree, this search is performed in $O(\max(Range) \times log|C_{\text{train}}|)$.
So, the total runtime complexity can be either $O(|Range| \times \log |C_{\text{train}}| + \max(Range) \times log|C_{\text{train}}|) = O((|Range| + \max(Range)) \times \log |C_{\text{train}}|) = \underline{\mathbf{O(\max(Range) \times \log |C_{\text{train}}|)}}$, or $O(|Range| \times l \times \log |C_{\text{train}}| + \max(Range) \times log|C_{\text{train}}|) = O((|Range| \times l + \max(Range)) \times \log |C_{\text{train}}|) = \underline{\mathbf{O(|Range| \times l \times \log |C_{\text{train}}|)}}$, depending on the similarity function $sim_f$ as explained above.

# 5 EVALUATION

We use five different classification tasks with geo-spatial coordinates (latitude and longitude) for evaluation. All presented experiments were conducted using 10-fold cross-validation, where we used each time 90% of the data as training data and 10% of the data as test data. In this section, we describe datasets and classification tasks. We evaluate expected accuracy measure w.r.t. different similarity functions. Finally, we analyze the accuracy of the ACKER classifier and discuss how to optimize the runtime of the ACKER reducing the *Range* and not decreasing the overall accuracy.

**Table 1: Description of the used datasets**

| Dataset | Geography | Classes |
|---|---|---|
| Twitter | Milan, Italy | Geographical items mentioned in tweets |
| WideNoise | Worldwide | Measured noise levels |
| SF Crimes | San Francisco, CA, USA | Types of crimes |

## 5.1 Data

We use three datasets that represent geo-spatial data from three different domains , see Table 1. We constructed five classification tasks, see Table 2, to evaluate the ACKER approach. Here, we shortly present each dataset and classification tasks:

*Milan Twitter Dataset.* The Twitter dataset[1] was collected in Milan between November 1, 2013 and December 31, 2013 for the Telecom Italia Big Data Challenge 2014. For each tweet we consider the entities extracted from the text by the dataTXT-NEX tool [1]: if an entity $x$ is mentioned in tweet, this means that this tweet belongs to the class $x$. For the classification tasks, `twitter8`, we consider tweets where one of the seven most popular local geographical items (e. g., “Milan Cathedral”, “Stadium San Siro”) is mentioned. We also added a general entity (“Italy”) in order to add some noise to the data. The second classification task, `twitter11`, consists of all tweets from `twitter8` and tweets with where one of three additional general entities is mentioned.

*WideNoise Dataset.* The WideNoise dataset [2] consists of worldwide noise measurements that were made by mobile app users [2]. We use measurements where GPS-based locations are available. The task is to predict a measured noise level. We test our approach with different number of classes: we divide noise levels into three (`widenoise3`) and six (`widenoise6`) classes. The classes in the classification tasks consist of the same number of samples (12,375 measurements per class in `widenoise3`, and 6,188 measurements per class in `widenoise6`).

*San Francisco Crimes Dataset.* The San Francisco Crimes dataset [3] consists of all crimes registered by the San Francisco Police Department during 2015. The crime data consists of the location of crimes (coordinates) and crime categories (e. g., “Burglary”, “Drug/-narcotic”). The task of the classifier is to predict the crime category, given the coordinates of that crime. We consider 16 most popular categories, each with over 1000 incidents in 2015.

[1] https://dandelion.eu/products/datatxt/
[2] http://kde.cs.uni-kassel.de/everyaware/dumps/widenoise
[3] https://data.sfgov.org/Public-Safety/SFPD-Incidents-Previous-Year-2015-/ritf-b9ki

**Table 2: Classification tasks constructed from the datasets**

| Name | Dataset | No. of Items | No. of Classes |
|---|---|---|---|
| `twitter8` | Twitter | 6,230 | 8 |
| `twitter11` | | 21,658 | 11 |
| `widenoise3` | WideNoise | 37,126 | 3 |
| `widenoise6` | | 37,126 | 6 |
| `crimessf` | SF Crimes | 149,401 | 16 |

Note, we did all of the following experiments for all five classification tasks for $k \in \{1..200\}$ and $l \in \{1..1000\}$. Due to space constraints, we do not outline redundant results: e.g., if only a plot for `widenoise6` is shown, it means, the plot for `widenoise3` has the same learnings; or if a plot for $k = 10$ is shown, plots for other values of $k$ have the same tendencies.

## 5.2 Expected Accuracy

We evaluate the expected accuracy measure – using the Pearson correlation and the ROC AUC. First, we aim to understand, whether this measure, based on the introduced similarity functions, reflects the real performance of standard kNN. We use the *lat_lon* similarity function as a baseline. The idea of this similarity function was inspired by [12]. We showed (see Equation 12) that their approach can be presented as a special case of the ACKER algorithm. Still, we use that approach as a baseline.

In this (and further experiments) we observe that results for `crimessf` classification task are different from other classification tasks. We assume, the reason for those differences is a skewed class distribution, (see Figure 2): there are few prevailing classes and many small classes. Over 28% of all items belong to the largest class. Further used datasets have more regular classes frequency.

We consider different values of $l$ for evaluation. The variable $l$ defines, how many similar cases should be found to choose $k$. This number can be chosen manually for given dataset. We discuss some strategies for choosing $l$ in the Section 5.3.

Expected accuracies (see Figure 3) generally have a high correlation with the observed accuracy. However, the correlation for small values of $l$ is rather unstable. For both Twitter and WideNoise datasets the expected accuracy based on the proposed similarity functions generally have a higher correlation with the observed accuracy than a *lat_lon*-based similarity functions. The `crimessf` classification task is the only exception, where the similarity function based on *lat_lon* seems to perform best. Generally, the correlations strongly depend on choice of $l$.

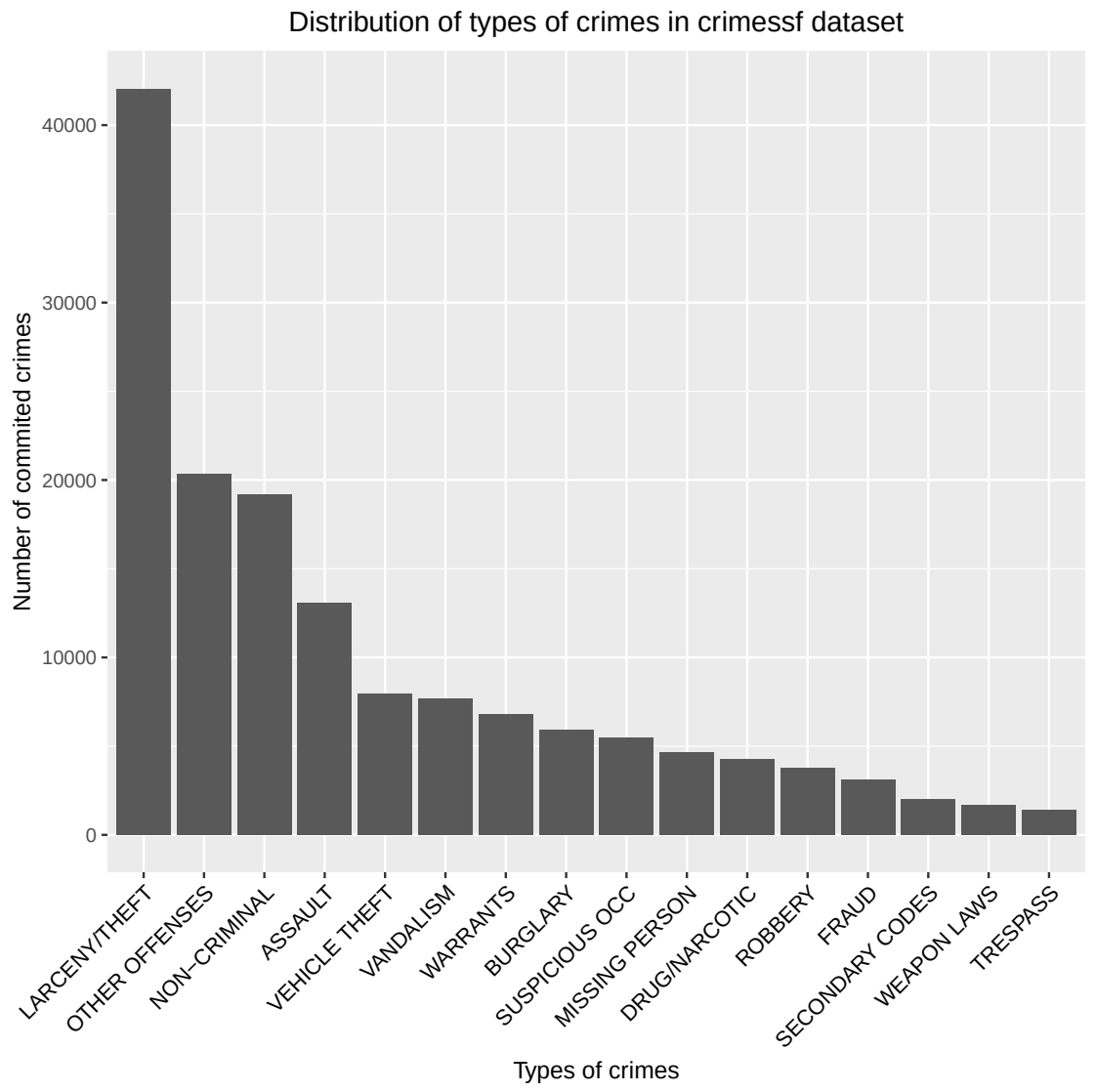

**Figure 2: Frequency of different crimes in `crimessf` classification task.**

Next, we compute the ROC AUC values for the same experiments to check if the expected accuracy measure is a good predictor of whether standard kNN classifies correctly. Figure 4 shows the ROC AUC scores values dependent on the different similarity functions and values of $l$ for the considered classification tasks. Overall, the similarity function performs as a good predictor for correctness of standard kNN classification. The expected accuracy performs better than a random predictor (ROC AUC $> 0.5$). moreover, expected accuracy, based on *lat_lon* similarity function, performs very well in terms of ROC AUC if the value of $l$ is low (it reaches almost 90% for the Twitter dataset).

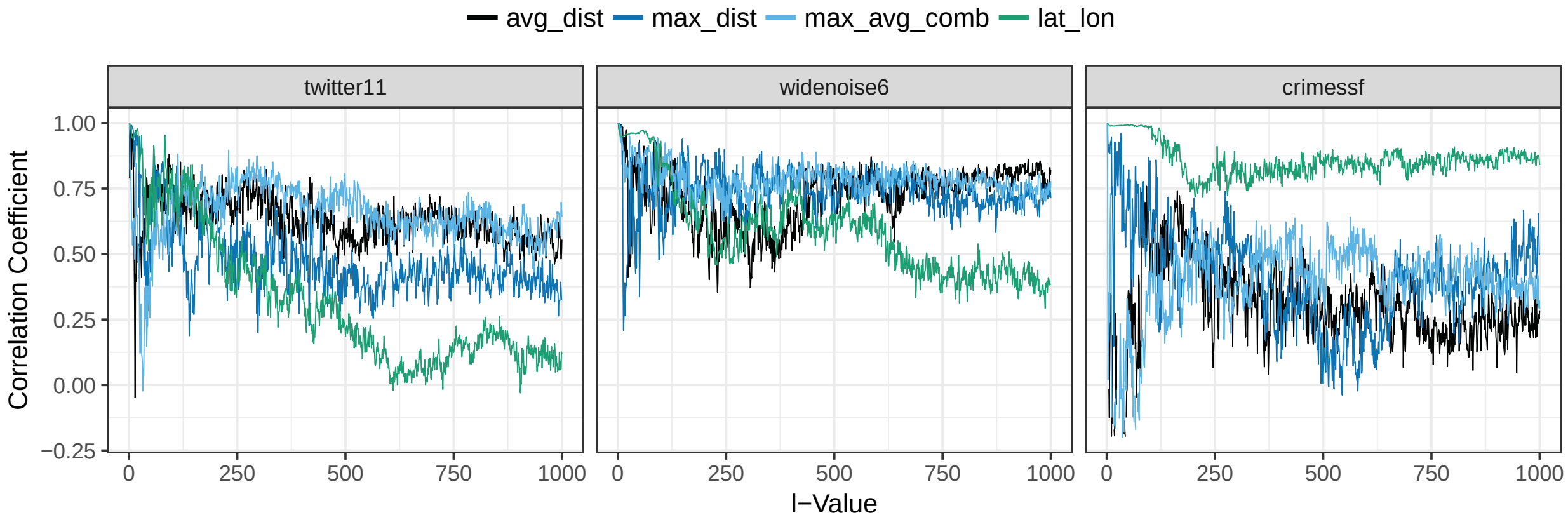

**Figure 3: Correlation between real and expected accuracy dependent on value of $l$**

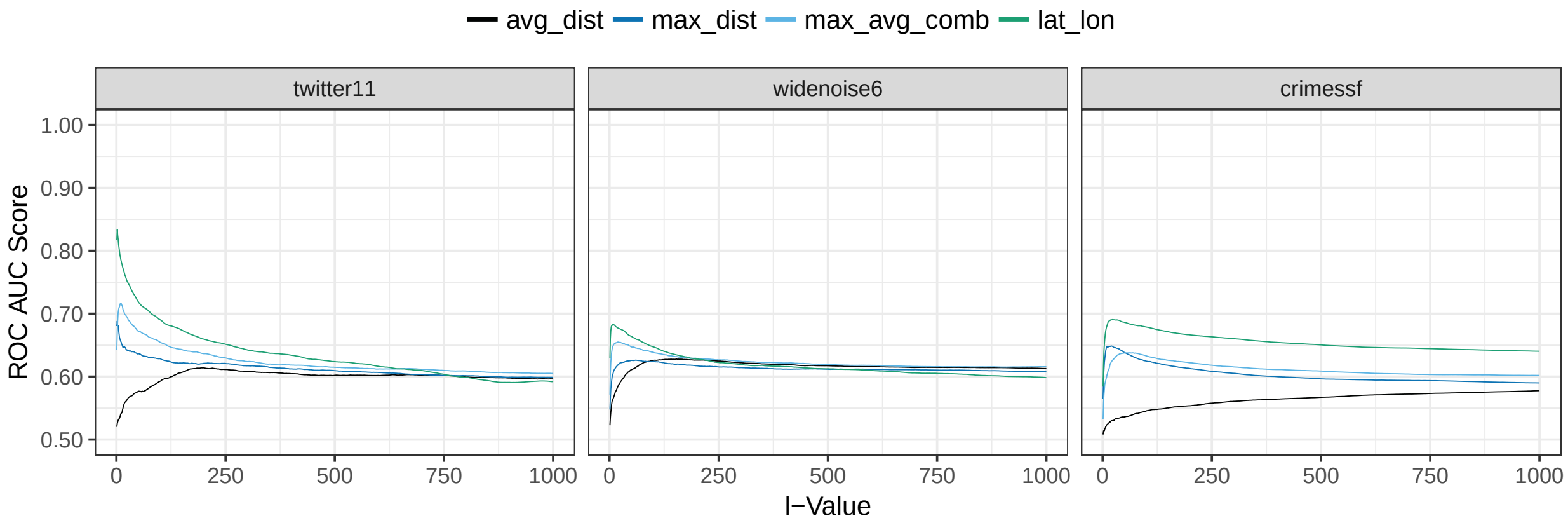

**Figure 4: ROC AUC scores of expected accuracy of standard kNN for $k = 200$.**

## 5.3 Adaptive kNN

Next, we discuss the performance of standard kNN and ACKER in terms of classification accuracy.

Figure 5 shows the mean accuracy and its standard deviation of standard kNN algorithm dependent on the choice of $k$. All classification tasks have different accuracy (note different Y-axis scales) as they have different number of classes, points and classes distribution.

For Twitter and WideNoise datasets, the standard kNN classifier performs best if $k$ is low. Its accuracy decreases if $k$ increases. In contrast, standard kNN with `crimessf` shows the worst performance with low $k$ (in fact, if $k$ is low, even choosing the most popular class performs better). Standard kNN performs best for `crimessf` with $k = 69$. For all classification tasks, the standard deviation is rather low, the results of standard kNN are stable.

We implemented the ACKER algorithm, as described in Listing 1. Figure 6 shows, how the accuracy for different datasets depends on the parameter $l$. Here, again, the results for `crimessf` classification task differ from the results for other classification tasks. First, consider the Twitter and WideNoise classification tasks: ACKER outperforms the standard kNN classifier, for the most $l$, e. g., the best possible accuracy for `twitter11` using standard kNN (with optimal $k$) is 77.3%, and the accuracy of ACKER (for *max_avg_comb* with optimal $l$) is: 82.5%. Also, the ACKER algorithm with introduced functions outperforms the baseline (we use adaptive kNN classification introduced in [12] as a baseline).

In contrast, for the `crimessf` classification task, only the ACKER with the *lat_lon*-based similarity function shows an improvement in terms of accuracy compared to the standard kNN. However, this improvement is marginal. This is the only classification task, where the proposed similarity functions performed poorer than the baseline.

For all classification tasks the choice of $l$ and similarity function is very important: e. g., for the classifier with the *avg_dist*-based similarity function of `twitter11` the minimal accuracy is 69.5% (with $l = 27$) and the maximal accuracy is 81.4% (with $l = 268$).

*How to choose an optimal similarity function and l?* In the previous section we considered two popular measures to estimate the relevance of expected accuracy – correlation coefficient and ROC AUC. The experiments results show that correlation coefficient is a much better indicator for choosing an optimal $l$ and similarity function. For the four classification tasks accuracy of the ACKER based on *lat_lon* similarity function performs poorer than ACKER based on the other functions; and vice versa for the `crimessf` classification task. The same observations can be found in correlation between expected accuracy and observed accuracy (s. Figure 3 and Figure 6): correlation coefficient with *lat_lon*-based expected accuracy is lower than with other expected accuracies for WideNoise and Twitter datasets and vice versa for San Francisco Crimes Dataset. In terms of ROC AUC, the *lat_lon*-based expected accuracy performed best for all classification tasks (see Figure 4). Considering $l$, it is necessary to notice that the ACKER algorithm performs better with higher values of $l$, the same as correlation coefficient is much stable with higher values of $l$. Again, ROC AUC values (see Figure 4) were lower with higher $l$ for all classification tasks. Thus, ROC AUC could not predict performance of similarity function and $l$.

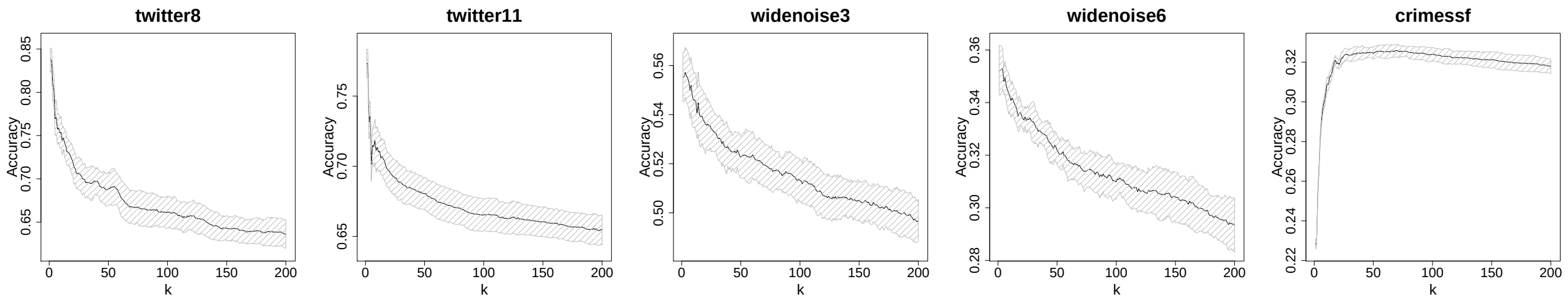

**Figure 5: 10-fold cross-validation mean accuracy and standard deviation of common kNN classifier dependent on value of $k$**

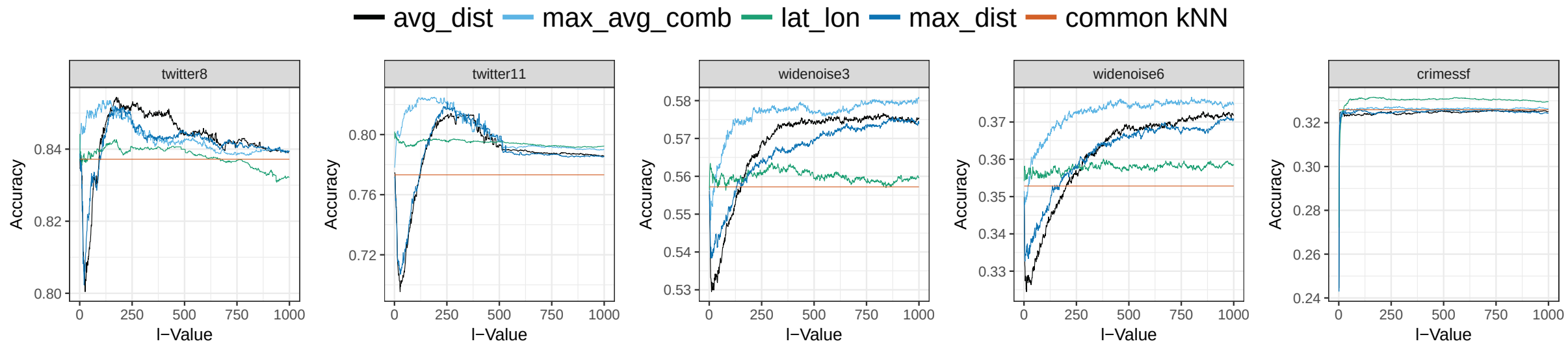

**Figure 6: Accuracy of the ACKER classifier dependent on the number of considered similar points $l$**

We conclude with some recommendations about the choice of the similarity function and $l$. One way to determine an optimal combination of similarity function and $l$ is to use the training dataset to calculate correlations between expected and real accuracy and analytically determine optimal similarity function and value of $l$.

It is also worth noticing that in many of our experiments the *max_avg_comb*-based ACKER classifier performs best: it has the best accuracy if an optimal $l$ is chosen in most scenarios.

## 5.4 Choice of Range of k

In the previous sections, we discussed the performance of ACKER for different similarity functions and numbers of considered similar points $l$. However, we did not discuss the choice of value of *Range*, see Listing 1. In all previous experiments we used $Range = \{1..200\}$. Since it is necessary that ACKER iterates over the *Range*, the runtime complexity of the ACKER algorithm is linearly dependent on number of values in *Range*, $|Range|$, see Section 4.4. Here we discuss, whether it is possible to reduce *Range* in order to speed up the runtime of ACKER.

In this experiment, for each classification task and each similarity function, we use fixed value of $l$, and particularly the value which performs best with $Range = \{1..200\}$, see Table 3. We evaluate, how the ACKER algorithm performs with different ranges of $k$, $Range = \{1..k_{max}\}$, where $k_{max} \in 1..200$.

As in previous experiments, we use the accuracy of the algorithm for evaluation – Figure 7 shows dependency of accuracy on $k_{max}$. Note, if $k_{max} = 1$, the algorithm is equivalent to the standard kNN with $k = 1$, and if $k = 200$, it is equivalent to the results discussed in Section 5.3.

For low values of $k_{max}$ even small increase of $k_{max}$ improves the accuracy of the ACKER classification for all classification tasks. Increasing $k_{max}$ if $k_{max} \geq 50$, does not increase the accuracy of ACKER for any classification task or similarity function, and even vice versa: e. g., consider *avg_dist*-based similarity function for `widenoise6`. It reaches maximum accuracy with $k_{max} = 41$.

**Table 3: Optimal values of $l$ (in terms of accuracy) for the considered classification tasks and similarity functions**

| Class. Task | *max_dist* | *avg_dist* | *max_avg_comb* | *lat_lon* |
|---|---|---|---|---|
| twitter8 | 169 | 175 | 113 | 1 |
| twitter11 | 251 | 268 | 191 | 1 |
| widenoise3 | 873 | 829 | 1000 | 298 |
| widenoise6 | 968 | 854 | 799 | 379 |
| crimessf | 734 | 742 | 290 | 166 |

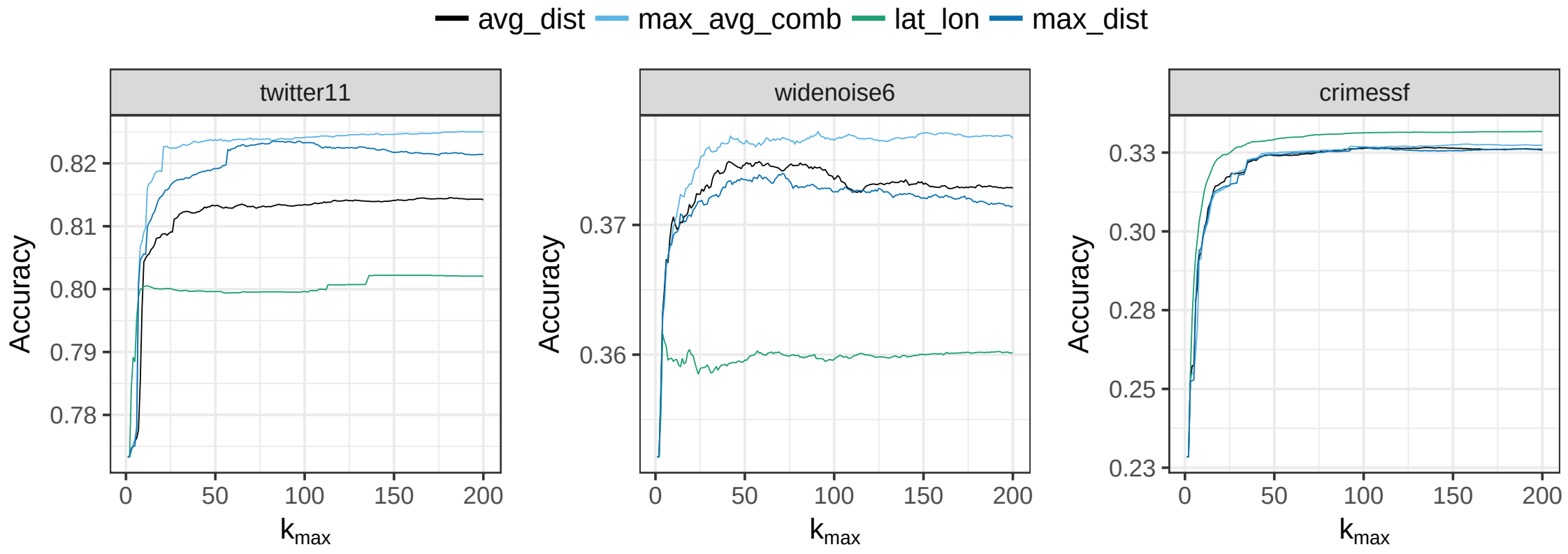

**Figure 7: The accuracy of ACKER classifier dependent on value of $k_{max}$ and different similarity functions**

These observations correspond with Figure 5: for the standard kNN algorithm the accuracy decreases if $k$ increases (Twitter and WideNoise), or does not change much after certain $k$ (San Francisco Crimes Data). The ACKER algorithm needs some number of $k$ values in *Range* to be able to "choose" from, but after certain value of $k_{max}$ the accuracy does not improve. For the choice of $k_{max}$, we suggest to use a visual elbow rule based on experiments as in Figure 5, as it is common for clustering algorithms [7].

## 6 CONCLUSIONS AND FUTURE WORK

In this paper, we presented an adaptive kNN classification algorithm – the ACKER algorithm. The basic idea of the algorithm is to choose $k$ for each item by maximizing its expected accuracy. Expected accuracy is computed based on similar points. We introduced three different similarity functions, and used method introduced by Sun and Huang [12] as the baseline. The evaluation was conducted using five different classification tasks, derived from three different datasets with geo-spatial data. The proposed ACKER algorithm based on three introduced similarity functions showed clear accuracy improvement compared to the standard kNN algorithm and to previously introduced adaptive kNN algorithms for the most classification tasks. The only dataset, for which ACKER could not improve the accuracy compared to standard kNN was San Francisco Crimes Dataset. One possible explanation, kNN approach can not perform well with this dataset, is a skewed class distribution.

Moreover, the expected accuracy demonstrated a fair ability to predict classification correctness, reaching ROC AUC of over 0.9 for some classification tasks and high correlation values with the observed accuracy. Further, we showed, it is not necessary to consider a large *Range* of $k$ values, as after some point, additional $k$s do not improve an accuracy. Particularly, reducing *Range* from 200 elements to 50 elements does not decrease accuracy, but can reduce runtime by four times (as it linearly depends on $max(Range)$, or $|Range|$).

The presented approach is able to improve classification using the topology of data. This is simple approach. Further, it is possible to consider the variability of considered phenomena for different problems. This is a possible way to extend the method. In this case, it would be necessary to make expected accuracy dependent on further factors.

Another promising topic for the future work is a more detailed investigation of different similarity functions and their influence on the expected accuracy. Also, we aim to apply the classifier on non-geo-spatial data in order to test its ability to work with multi-dimensional data. We discussed some heuristics to choose the right parameters for ACKER. However, more formal approaches for the choice of the input parameters of ACKER could be very useful for future applications.